\documentclass[letterpaper, 10 pt, conference]{ieeeconf}
\usepackage{amsmath,amsfonts}
\usepackage{array}
\usepackage{enumerate}
\usepackage[utf8]{inputenc}
\usepackage[T1]{fontenc}
\usepackage{amsfonts}
\usepackage{amssymb}
\IEEEoverridecommandlockouts 
\usepackage{amsmath}  
\usepackage{breqn}
\usepackage[caption=false]{subfig}
\usepackage{textcomp}
\usepackage{stfloats}
\usepackage{url}
\usepackage{verbatim}
\usepackage{graphicx}
\usepackage{cite}
\usepackage[linesnumbered, ruled]{algorithm2e} 
\usepackage{hyperref}
\usepackage{diagbox}
\usepackage{amsfonts}
\usepackage{makecell}
\usepackage{adjustbox}
\usepackage{float}
\usepackage{color}
\usepackage{booktabs}

\title{\LARGE \bf
Unifying Large Language Model and Deep Reinforcement Learning for Human-in-Loop Interactive Socially-aware Navigation
}

\author{Weizheng Wang$^{1}$, Ike Obi$^{1}$, Aniket Bera$^{2}$, and Byung-Cheol Min$^{1}$ 
\thanks{$^{1}$SMART Laboratory, Department of Computer and Information Technology, Purdue University, West Lafayette, IN, USA. {\tt\small{[wang5716, obii,minb]@purdue.edu}.}
}
\thanks{$^{2}$Department of Computer Science, Purdue University, West Lafayette, IN, USA. {\tt\small{aniketbera@purdue.edu}.}
}}

\begin{document}

\maketitle

\begin{abstract} Navigating human-filled spaces is crucial for the interactive social robots to support advanced services, such as cooperative carrying, which enables service provision in complex and crowded environments while adapting behavior based on real-time human language commands or feedback. However, existing social robot navigation planners face two major challenges: managing real-time user inputs and ensuring socially compliant behaviors in unfamiliar, zero-shot environments. In response, we introduce SALM, an interactive, human-in-loop Socially-Aware navigation Large Language Model framework that dynamically integrates deep reinforcement learning (DRL) with large language model (LLM) capabilities. SALM leverages contextual semantic understanding from real-time human-robot interactions to convert high-level user commands into precise, low-level control actions. A high-level LLM module parses user input, guiding the simultaneous generation of navigation commands by both a large language navigation model (LNM) and a DRL-based navigation model (RLNM). A memory mechanism archives temporal data for continuous refinement, while a multi-step graph-of-thoughts inference-based large language feedback model adaptively fuses the strengths of both planning approaches. Experimental evaluations demonstrate that SALM not only enhances navigational precision in crowded, dynamic environments but also significantly improves system adaptability, offering tailored behaviors that align with individual user preferences and real-time feedback. More details and videos about this work are available at: \url{https://sites.google.com/view/navi-salm}.
\end{abstract}

\section{Introduction}

Navigating in human-filled spaces is crucial for social robots to support various of advanced services \cite{tro-hri1, tro-hri2, tro-hri3}, such as human-robot cooperative carrying. The socially-aware navigation (SAN) task still faces two primary challenges: managing highly volatile, real-time user requests or sentiments, and ensuring socially compliant navigation behaviors in zero-shot unfamiliar environments. Recent developments in robotics and artificial intelligence have led to approaches that address these challenges for deploying social robots in public spaces \cite{san-tro-1 ,san-ijrr-1}. These solutions draw upon insights from machine learning \cite{wang2023navistar}, sociology \cite{mavrogiannis2019multi}, analytical mechanics \cite{navidiff, mavrogiannis2021hamiltonian}, algebra and geometry \cite{trautman2015robot}, among other fields, to achieve robust and socially-aware navigation.

\begin{figure}[!t]
\centering
\includegraphics[width=1\columnwidth]{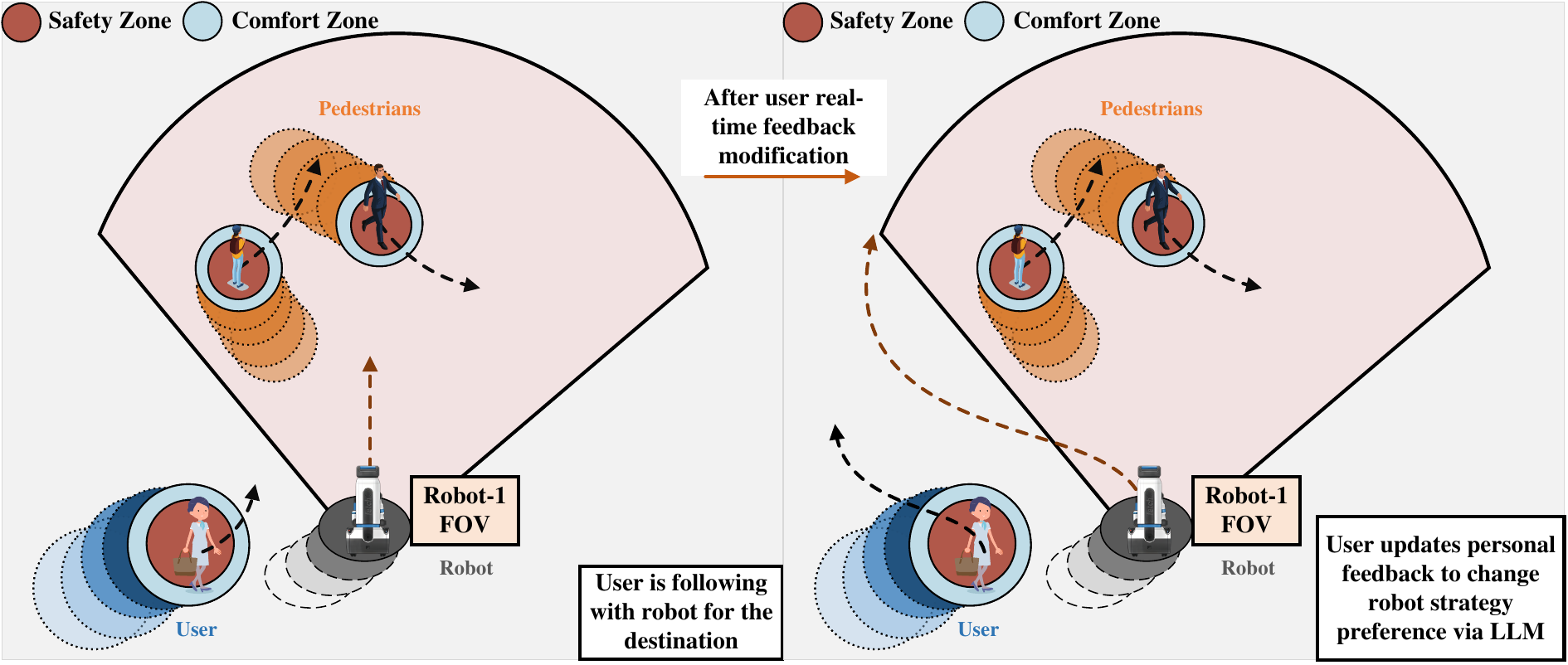}
\vspace{-10pt}
\caption{An illustration of human-in-the-loop interactive social navigation: integrating DRL and LLM enables the social robot to adaptively respond to environmental changes such as task target and preferred social distance from real-time human feedback.}
\vspace{-20pt}
\label{fig:F1}
\end{figure}

However, both existing learning-based and conventional approaches demonstrate limited adaptability and transferability when addressing real-time response requirements in unfamiliar or zero-shot environments. In response, we introduce an interactive \textbf{S}ocially \textbf{A}ware navigation large \textbf{L}anguage \textbf{M}odel (SALM) that dynamically adapts to environmental changes by leveraging the potential of large language models (LLMs) for understanding contextual semantics in human-robot interaction (HRI) based on users' real-time commands and feedback. For example, users can adjust the robot’s configuration or behavioral style according to their personal preferences, which a flexibility that current state-of-the-art (SOTA) planners often lack due to their fixed, converged policy parameters, as shown in Fig. \ref{fig:F1}.

% Hence, SALM interprets real-time human feedback inference via an LLM as high-level global information to guide low-level actions execution. This interactive framework not only enhances user experience but also boosts performance,

On the other hand, despite the widespread applications of LLMs in robotics \cite{shah2023navigation}, HRI \cite{li2023interactive}, and logical reasoning \cite{CoT,GoT}, the deployment of LLM-based social navigation planners remains constrained. These limitations arise from gaps in real-time navigation information and biases inherent in standard pre-training datasets. Moreover, many current LLM-based navigation systems \cite{shah2023navigation, shah2023lm} generate macro-action functions rather than low-level motion control commands, largely due to LLMs' insensitivity to continuous numerical values. To address these challenges and improve navigational precision, SALM integrates a textual embedding encoder with a tailored social navigation prompt. This combination effectively translates environmental data into direct low-level navigation commands, ensuring consistent inference between environmental semantic understanding and path planning.

We propose an interactive, embodied intelligent social navigation system that is tailored to each user's personal preferences and real-time feedback. Our approach seamlessly integrates a SOTA DRL navigation planner with LLM-based techniques into a unified navigation framework. This hybrid methodology addresses the limitations of current LLM-based planners, particularly their challenges in handling complex dynamic optimization problems in crowded environments, while also enhancing the transferability of RL-based planners. The main contributions of this paper are as follows:
\begin{itemize}
    \item We propose SALM, a human-in-loop interactive \textbf{S}ocially-\textbf{A}ware navigation large \textbf{L}anguage \textbf{M}odel framework, incorporating both LLMs and deep reinforcement learning (DRL). SALM harnesses users' requests, preferences, and real-time feedback from human language to craft tailored tasks for social robots, functioning as an engaging, interactive assistant.
    \item SALM executes interactive tasks using a high-level user parsing LLM module that directs the generation of low-level control behaviors from both the large language navigation model (LNM) and the reinforcement learning navigation model (RLNM). This approach facilitates furthermore contextual semantic understanding to produce socially compliant robot behaviors. Additionally, the memory mechanism of LNM archives temporal data, providing long-term evaluations and feedback to continuously refine the DRL-based and LLM-based social planners.
    \item SALM adaptively integrates the unique strengths of both the LNM and the RLNM by incorporating a large language feedback model (LFM) scorer, utilizing multi-step Graph-of-Thoughts (GoT) inference technology.
    \item SALM demonstrates robust and promising exhibitions of socially compliant behaviors in various experiments.
\end{itemize}

\section{Background}

%SALM target -> (1)large model in dynamic environment; (2)and interactive framework; (3) AGI personal social robot assistant\\
%Challenges: 
%1. perception info -> text (input difference); develop large model into navigation task; improve the adaptability and robustness of pre-trained RL policies so that align to real-time user requirments (LNM)
%2. complex dynamic optimization problem -> too many conditions and objectives -> current only LLM can not generate a high accuracy (RLNM)
%3. limited inference ability (LFM+GoT)

\noindent\textbf{LLM-driven Navigation:} Inspired by the promising potential of LLM across diverse applications, recent breakthroughs powered by extensive computational resources and advanced machine learning techniques have catalyzed significant research in robotics \cite{navidiff, shah2023lm} and HRI \cite{tidybot, li2023interactive}. The robust inference capabilities of LLM are essential for interactive social robots, substantially enhancing both the transferability and generalizability of pre-trained configurations to zero-shot environments. In this work, we leverage a high-level LLM module to parse user requirements as global guidance for low-level execution LLM as LNM.

% For instance, \cite{tidybot} designs an LLM-driven mobile manipulator that offers services or infers user preferences from human language requests. 

Recent LLM-based social navigation planners \cite{sun2024trustnavgpt, song2024vlm, huang2024drivlme, zu2024language} have demonstrated impressive capabilities in inferring contextual semantics and planning paths in human-populated environments. However, these approaches typically generate only semantic goals (e.g., go to position $p_A$) rather than direct low-level control signals (e.g., [$v_x, v_y$]) required for precise robot motion. This lack of low-level execution impedes both the accuracy and continuity of navigation. To address this limitation, SALM abstracts environmental features of social navigation into textual information for effectively capturing dynamic HRI nuances via the LNM block, enabling the direct generation of low-level robot velocity actions, as shown in Fig.~\ref{fig:F3}.

% Moreover, considering the ephemeral limitations of LLM inference, particularly due to potential probabilistic illusions in LLM reasoning and the insensitivity of sequential floating-point numbers in LLM, the RLNM is introduced to maintain baseline performance. Subsequently, the LNM and RLNM are adaptively integrated by the LFM, in which LFM evaluates and scores the actions pair from LNM and RLNM to final action selection. Additionally, to further improve the inference ability, the chain-of-thoughts \cite{CoT, GoT, feng2023towards} technology has been developed to generate intermediate steps of inference process that chain-type construction provides more generative information and an adjustable method. Thus, the graph inference structure is implemented in LFM for logical and detailed action evaluation, as shown in Fig.\ref{fig:F4}.

Moreover, to address inherent limitations in LLM inference, such as potential probabilistic hallucinations in reasoning and the insensitivity to sequential floating-point numbers, the RLNM is introduced to ensure robust baseline performance. The LNM and RLNM are then adaptively integrated via the LFM, which evaluates and scores action pairs from both models to select the final action. To further enhance inference capabilities, SALM employs chain-of-thoughts \cite{CoT, GoT, feng2023towards} technology, which generates intermediate reasoning steps, offering richer generative insights and increased flexibility. Consequently, a graph inference structure is implemented within the LFM for logical and detailed action evaluation, as illustrated in Fig.\ref{fig:F4}.

\noindent\textbf{Socially Aware Robot Navigation:}
% After early applications of robotics in social navigation society, such as MINERVA \cite{thrun1999minerva}, socially aware robot navigation tasks are primarily conducted via decoupled and coupled strategies \cite{mavrogiannis2023core}. Decoupled approaches infer pedestrian motion intents and patterns to construct potential safe areas for planing. However, the separation of modeling and planning often overlooks potential cooperation, leading to the establishment of limited feasible spaces, known as the freezing robot problem \cite{freezing}, particularly  with the increasing presence of humans. Alternatively, coupled approaches encode potential cooperation into navigation inference to address unwarranted ignorance. 
Building on early applications of robotics in social navigation, such as MINERVA \cite{thrun1999minerva}, current research in socially aware robot navigation typically follows either decoupled or coupled strategies \cite{mavrogiannis2023core}. In decoupled approaches, pedestrian motion intents and patterns are first inferred to delineate potential safe areas for planning. However, this separation between modeling and planning can neglect cooperative behaviors, often resulting in overly restrictive feasible spaces and triggering the "freezing robot problem" \cite{freezing} that becomes more pronounced with higher human density. In contrast, coupled approaches integrate cooperative interactions directly into the navigation inference process, thereby mitigating these limitations.

% On the other hand, explicitly coupled approaches are implemented through game theoretic planning \cite{schwarting2021stochastic}, Gaussian processes \cite{trautman2020real}, and topology analysis \cite{mavrogiannis2019multi}. However, the challenge of optimization in highly dynamic environments restricts the further deployment of conventional approaches such as ORCA \cite{ORCA} and explicitly coupled approaches, especially with the increasing complexity of the environment. Recently, the paradigm of cooperative collision avoidance has facilitated a set of promising works \cite{navidiff, wang2023navistar, liu2023intention}, which implicitly approximate human-like navigation awareness and insight through advanced neural networks to encode potential human-robot cooperation and compliance with social norms into robot behavior. These neural networks are then trained using DRLs to iterate through different situations. For instance, efforts have been made in the development of neural network technologies for social navigation, such as attention mechanisms \cite{chen2019crowd}, graph attention \cite{liu2023intention}, transformers \cite{wang2023navistar}, and diffusion \cite{navidiff}. 

Alternatively, explicitly coupled approaches are implemented through game-theoretic planning \cite{schwarting2021stochastic}, Gaussian processes \cite{trautman2020real}, and topology analysis \cite{mavrogiannis2019multi}. However, the optimization challenges posed by highly dynamic environments hinder the practical deployment of conventional methods like ORCA \cite{ORCA} and other explicitly coupled techniques, especially as environmental complexity increases. Recently, the cooperative collision avoidance paradigm has spurred promising research that implicitly captures human-like navigation awareness. These approaches leverage advanced neural networks to encode potential human-robot cooperation and compliance with social norms into robot behavior. Trained using deep reinforcement learning (DRL), these networks iterate through diverse scenarios, with recent advancements incorporating technologies such as attention mechanisms \cite{chen2019crowd}, graph attention networks \cite{liu2023intention}, transformers \cite{wang2023navistar}, and diffusion models \cite{navidiff}.

Despite the utilization of DRL to evaluate underlying human–robot interactions and pedestrian intents for cooperative collision avoidance, human preferences remain inadequately represented. Thus, \cite{SunM-RSS-21} incorporates the high-order uncertainty of human movements by modeling pedestrian preference distributions within a social navigation planner based on variational analysis. Furthermore, the SAN task has been advanced by directly involving human intelligence through reinforcement learning from human feedback, as demonstrated in \cite{wang2023navistar, navidiff}, where human expectations and social norms are integrated into the training policy via a reward neural network. More recently, extensions to the learning-based SAN planner have been proposed for multi-robot scenarios \cite{wang2023multi, hypersamarl}, further broadening the scope of cooperative navigation strategies.

% Despite aforementioned neural networks being utilized to evaluate underlying human-robot interaction and pedestrian intents for cooperative collision avoidance, human preferences are still not well represented. \cite{SunM-RSS-21} incorporates the high-order uncertainty of human movements as pedestrian preference distributions into the social navigation planner based on variational analysis. Furthermore, the SAN task is also motivated by the direct involvement of human intelligence through RLHF in \cite{wang2023navistar, navidiff}, where human expectations and social norms are studied and embedded by a reward neural network to train the policy. More currently, \cite{wang2023multi, hypersamarl} extends the learning-based planner in multi-robot scenarios. 

While learning-based approaches typically perform well on benchmarks, DRL policies often struggle in zero-shot unfamiliar environments, leading to performance degradation and a disconnect from real-time human sentiment. For example, a DRL-based policy trained with a fixed social distance parameter (d = 0.4) may fail to adapt when a user interacts with the robot in ways that require different proximities such as approaching closely to hand over a package (d = 0) or keeping a greater distance (d = 1.5). To address this challenge, we integrate the large language navigation model (LNM) and the large language feedback model (LFM) into SALM. This integration harnesses the inference capabilities of large language models to process real-time human feedback and generate precise low-level motion commands, thereby enhancing the adaptability and robustness of DRL-based planners.

\begin{figure*}[!t]
\centering
\includegraphics[width=0.95\linewidth]{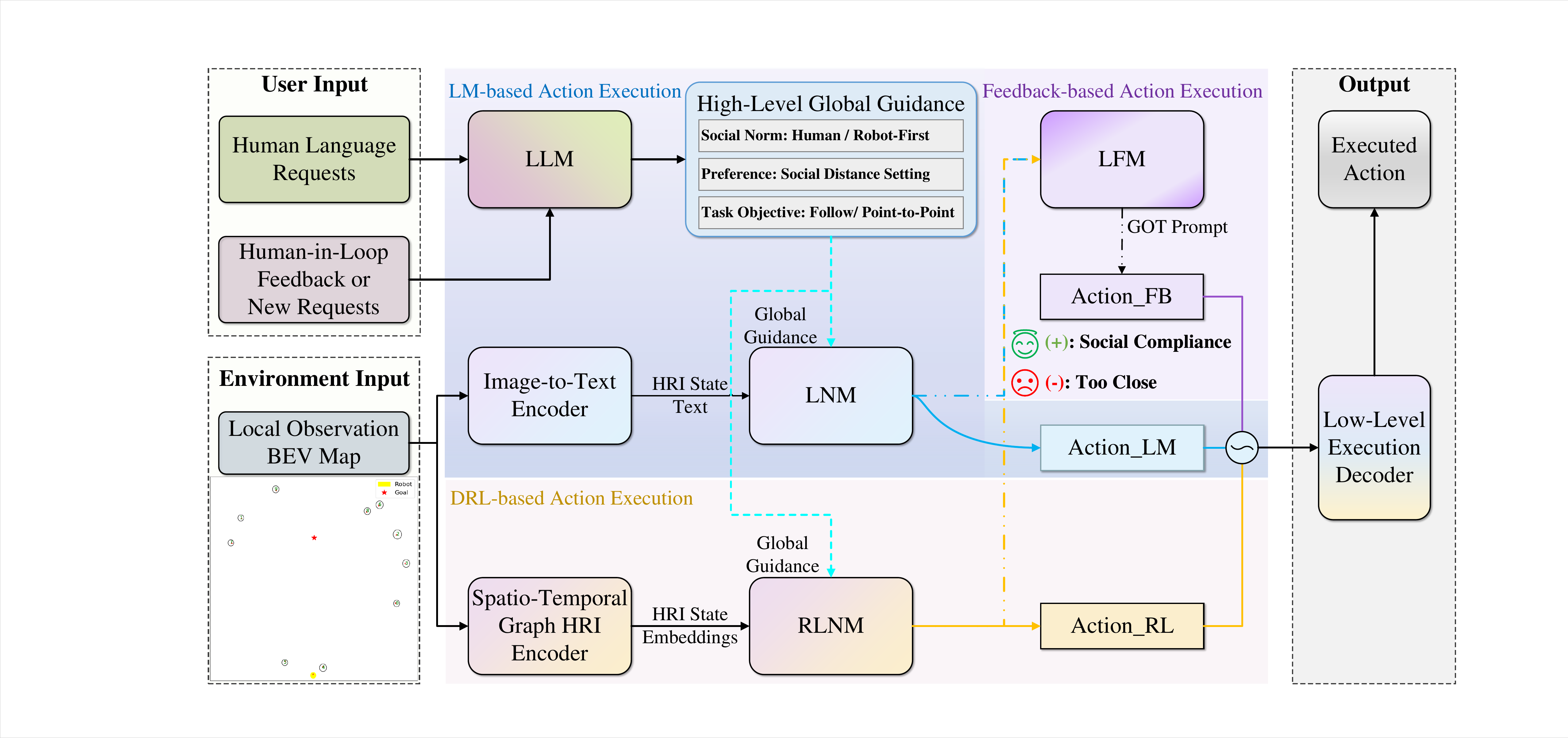}
\vspace{-10pt}
\caption{SALM architecture: SALM is implemented as a human-in-loop interactive social robot navigation framework, which executes human commands based on LM-based planner, feedback-based planner, and DRL-based planner incorporating. Firstly, users' requests or real-time feedback are processed or replanned to high-level task guidance for three action executors via LLM. Then, the image-to-text encoder and spatio-temporal graph HRI encoder convert robot local observation information to features as LNM and RLNM input, which generate RL-based action, LM-based action, and feedback-based action. Lastly, the above three actions are adaptively fused by a low-level execution decoder as the robot behavior output of SALM.}
\vspace{-15pt}
\label{fig:F2}
\end{figure*}

\vspace{-5pt}
\section{Preliminary}

SALM is an interactive social navigation framework that understands user commands (e.g., ``Pick up my bag to me") and personal preferences (such as the user needing a larger privacy area with the robot) into a set of high-level instructions as global guidance from human language input. Herein, SALM blends a high-level and low-level execution system in which task objectives, such as point-to-point (P2P), human-follow (HF), user preferences (privacy distance), and social norm property (whether to wait each pedestrian), are composed into high-level global guidance for low-level robot action generation.

Subsequently, the pre-trained LNM and RLNM generate low-level robot actions with respect to textual or featurized HRI state presentations and the above global guidance. Moreover, the global guidance information can be modified or updated by real-time human feedback as well, such as when a user adds personal feelings or changes task objectives, prompting instructions to be replanned. Lastly, an additional feedback and memory mechanism is adapted to adjust LNM \& RLNM, incorporating behaviors from past trajectories. The real-time evaluation and feedback mechanism provides an adaptive heuristic to take advantage of. For instance, the DRL-based action execution model maintains a lower bound when the LM-based model encounters explicit mistakes or dangers, thus providing fundamental performance as the social navigation benchmark. On the other hand, the LM-based action execution model incorporates user personal modifications from user language to improve the robustness of DRL-based action execution model.

Herein, the interactive social robot navigation problem is formulated as a Dec-POSMDP (decentralized particularly observable semi-Markov decision process) problem based on \cite{wang2023multi}, characterized by the tuple $\langle \mathcal{S},\mathcal{U},\mathcal{A},\Omega,\mathcal{O},\mathcal{P} ,\mathcal R, \mathbf{R}, \mathcal{C}, \mathcal S_0, \gamma,\rm N, \Upsilon \rangle$. Here, $\rm \mathbf{{s}}_{\rm{t}} = [\rm \mathbf{s}^{r}_{\rm{t}},\rm \mathbf{s}^{h_1}_{\rm{t}}, \cdots, \rm \mathbf{s}^{h_N}_{\rm{t}}] \in \mathcal{S}$ denotes the fully observable and unobserved states of the robot and humans at the $\rm t$-th timestep, belong to the state space, with the observable state denoted by $ {\rm \mathbf{s}^{o}_{\rm{t}}} = [ p_{\rm x}, p_{\rm y}, v_{\rm{x}}, v_{\rm{y}},\rho]$ covering individual position, velocity, and radius information that can be estimated by the robot. Accordingly, pedestrians' personal preferences and intent goals remain unobserved by robots, represented as $\mathbf{s}^{\rm{uo}}_{\rm{t}}=[ {g}_{\rm{x}}, {g}_{\rm{y}},v_{\rm{pref}}]$. Moreover, $\mathrm{{u}_{t_{}}} \in \mathcal{U}$ represents robot macro-action (MA), such as waypoint locations or robot operations, which are adaptive to real-time user requirements or feedback, while robot local-action (LA) are denoted by $\mathrm {a_t} = [v_x, v_y] \subseteq \mathcal{A}$, representing velocity. $\mathcal{O} \rm {(s^{o}|(s,a))}$ denotes the observation probability of the robot in the observation space $\Omega$, and $\mathcal{P}$ represents the state transition probability. $\mathcal R,~\text{and}~ \mathbf{R}$ represent the MA reward and LA hand-crafted reward functions separately. The MA reward is generated by the following objective: ${\mathcal{R}}(\mathrm{{s},{u}}) = \mathop{\arg\max} \mathbb{E} [\textstyle\sum_{\rm t=0}^{\rm T} \gamma^{\rm t} {\mathbf{R}}({\rm s}_{\rm t},{\rm a}_{\rm t})|{\rm a}_{\rm t} \sim {\rm u}]$. Additionally, $\mathcal{C}$ represents the conditional function, which can be updated by user language comments from LLM, $\mathcal{S}_{0}$ is the initial distribution, $\rm N$ is the number of pedestrians, $\gamma \in [0,1]$ is a discount factor, and $\Upsilon$ denotes the LLM global guidance wherein task objectives, user preferences, and robot properties are involved.

The interactive social navigation task statement can be viewed as a condition of a single robot from the multi-social robot navigation task definition \cite{wang2023multi}. The simulation environment also adheres to the same robot kinematic and dynamic configurations \cite{navidiff, wang2023navistar}. For further definitions and theorems of Dec-POSMDP, refer to  \cite{POSMDP, wang2023multi}.

%\vspace{-8pt}
\section{Methodology}
%\vspace{-4pt}
SALM leverages multiple LLM-based large models and a DRL-based model to provide interactive social robotic services under the dynamically transferable environment that is different with the training configuration of DRL-based LNM robot strategy, with respect to users' requirements or feedback. In this framework, the LLM-based navigation large model (LNM) and DRL-based approach (RLNM) are adaptively incorporated with an LLM-based evaluation model (LFM), as shown in Fig~\ref{fig:F2}.

% SALM leverages multiple large language model (LLM)-based systems alongside a deep reinforcement learning (DRL) model to deliver interactive social robotic services in dynamic environments that differ from the original DRL training configuration. This framework adapts to users' requirements and feedback by seamlessly integrating an LLM-based navigation model (LNM) with a DRL-based approach (RLNM) and an LLM-based evaluation model (LFM), as illustrated in Fig.~\ref{fig:F2}.

%LNM (Language Navigation Model): Executes social navigation behaviors via an LLM-based inference block.

%RLNM (Reinforcement Learning Navigation Model): A pre-trained policy network executes social navigation motions with respect to the maximum expected return objective in Dec-POSMDP.

%LFM (Language Feedback Model): A block based on LLM is developed to evaluate the output actions of LNM and RLNM and to estimate the related weight parameters of such actions.

\begin{figure*}[!t]
\centering
\includegraphics[width=0.95\linewidth]{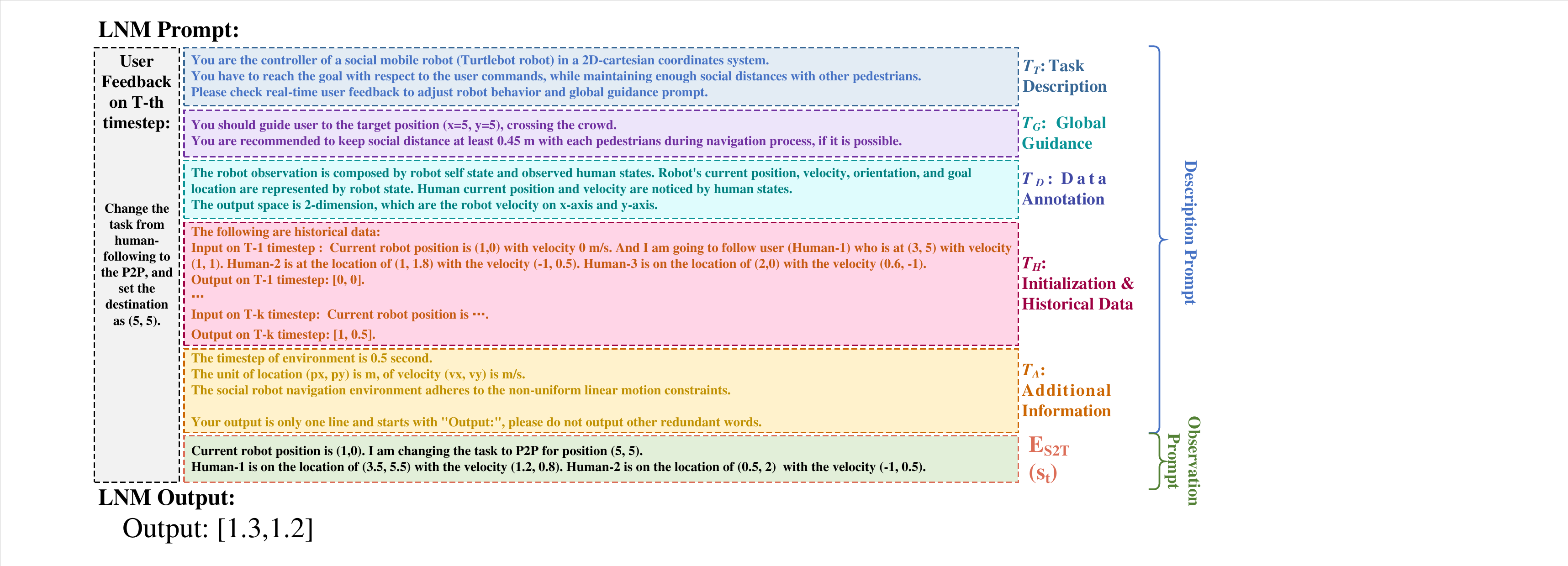}
\vspace{-5pt}
\caption{An illustration of large language navigation model (LNM): The prompt engineering of LNM comprises task description, global guidance, data annotation, initialization, historical data, additional information, and encoded state to directly generate low-level robot actions $[v_x, v_y]$.}
\vspace{-10pt}
\label{fig:F3}
\end{figure*}

\subsection{Human-in-Loop Interactive Mechanism}
 
SALM drives social robots through high-level guidance and low-level execution incorporation strategy. Firstly, LLM handles user language input to capture semantic features for global guidance generation. We define two typical social robot tasks. In the HF task, the robot's target is updated by the user's real-time location, and the robot must ensure that the user is within its field of view (FOV) at a comfortable social distance. The basic P2P task involves simply assigning a new target to robot. Moreover, social norm attributes are considered, such as pedestrian-first or robot-first, where the robot will come to a full stop when a pedestrian appears within a fixed distance ($d_s$) area under the condition of pedestrian-first. 

%Additionally, user personal preference attributes are collected to select different styles of DRL policy networks in RLNM. Here, we trained three pre-trained policy networks with preferences for large, moderate, and minimal social distance. 

Subsequently, global guidance $T_G$ is further employed in the following low-level execution blocks: LNM, RLNM, and LFM, where global guidance is described in the prompt engineering of LNM and LFM to supervise LM-based and FB-based action execution. RLNM encodes the target and personal attributes into $\Upsilon$, which can modify the conditional function $\mathcal{C}$.

Additionally, the LLM block can modify the global guidance or replan new global information based on real-time user feedback or new requests. The human-in-loop interactive mechanism enhances the robustness and flexibility of SALM, allowing users to adjust robot behaviors based on their feelings during the real-time execution process. For example, if the robot is too closed to pedestrians, users can provide personal feedback to modify the robot's social distance to a larger value. Moreover, the employment of LNM address the generalization and transfer issues as well, using the robust commonsense inference ability.

\subsection{Large Language Navigation Model}
LNM adapts LLM's supervising ability of context semantic inference to drive a social robot as a low-level motion controller in a human-filled environment. Due to the current requirement of textual information input by LLM, the perception information of the social robot (such as the position of pedestrians) has to be converted into textual information via an image-to-text encoder. Here, the image-to-text encoder translates the robot's observation state into text descriptions, which mainly include the location and velocity of the robot and observed pedestrians, as well as other features such as orientation and personal radius. However, simply feeding the robot's observation and action pairs into LLM cannot produce a robust controlling sequence due to the insensitivity and misunderstanding of LLM regarding a set of numeric values. To enhance the LLM's inference ability on temporal series, SALM also implements the prompt engineering $T_{HRI}$, which consists of the following parts: task description $T_T$, global guidance $T_G$, data annotation $T_D$, initialization and historical data $T_H$, and additional information $T_A$, as shown in Fig.~\ref{fig:F3}.

Firstly, task description $T_T$ is a paragraph that explains the environment configuration and robot properties. The global guidance $T_G$ notes immediate task objectives, user personal preference requirements, and social norm conditions, which are abstracted from the output of the first LLM block. The third subsection is data annotation $T_D$, which specifies the implications and data formulation of inputs and outputs. The demonstration data or executed actions are saved into initialization and historical data $T_H$. Finally, additional information $T_A$ provides supplementary information.

In the execution process, the initialization textual information from demonstrations is saved in first several time steps after the robot receives user commands, and then the demonstration data will be replaced with historical data by a memory mechanism.
\begin{equation}
\begin{aligned}
&\mathrm{T}_{\rm HRI} = \{\mathbf{E}_{\rm S2T}(\mathbf{s}_{\rm t}),~LLM(Requests)\}\\
&\mathbf{a}^{\rm LM} = LNM(\mathrm{T}_{\rm HRI}) \\
\end{aligned}
\end{equation}
\noindent where $\mathbf{E}_{\rm S2T}(\mathbf{s}_{\rm t})$ denotes the textual representation of robot state $\mathbf{s}_{\rm t}$, which is a language description of robot observation information. $LLM(Requests)$ is the high-level guidance from global LLM, representing real-time user commands.

%LNM is initialized by the data collections and demonstrations from existing SOTA social navigation approaches. The observation \& action pairs from demonstrations are translated to text description firstly, and then are added into LNM prompt which perform benchmarking performance of social navigation behavior inference for LNM. In the execution produce, the initialization textual information are saved in first several timesteps after robot receives user commands, and then the demonstration data will be replaced with historical execution data by a memory mechanism.  The data transformation is defined as follows:

%Additionally, all the continues data are normalized and discretized into a non-negative range to improve the data effectiveness, because current LLM has struggle with the negative and float numbers input.

\begin{figure*}[!t]
\centering
\includegraphics[width=0.95\linewidth]{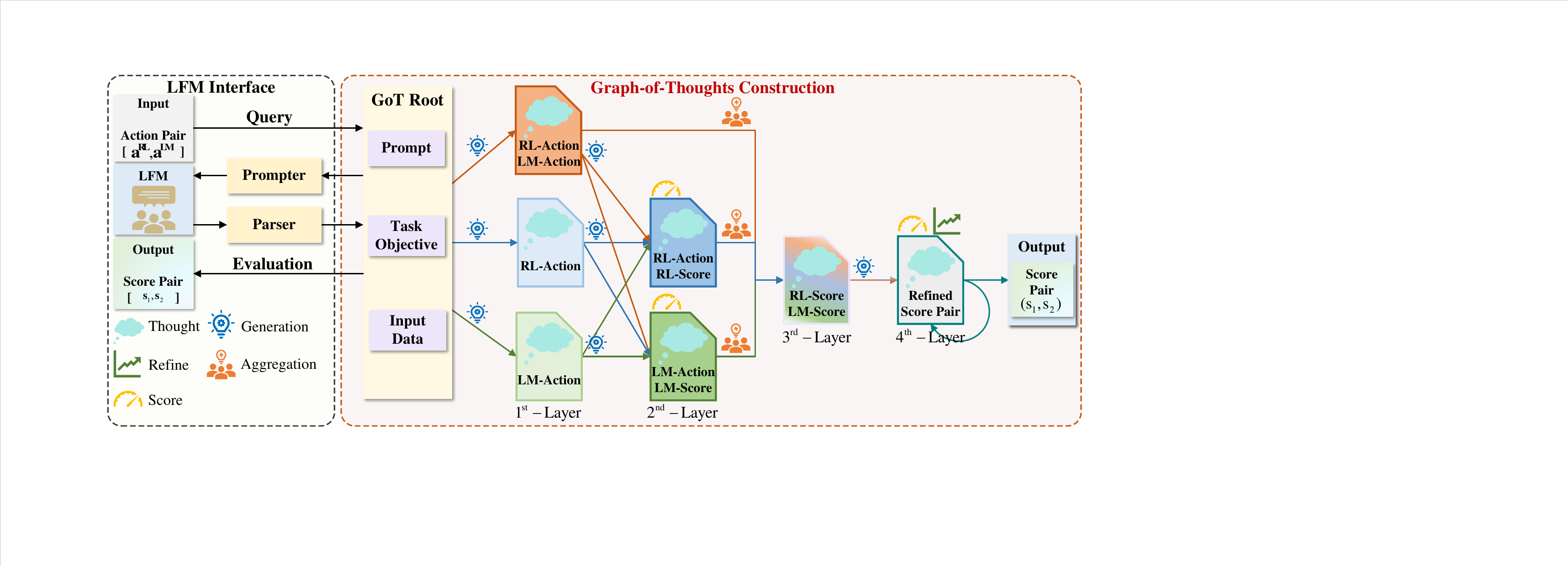}
\vspace{-5pt}
\caption{LFM framework: LFM reconciles the output from LNM $\mathrm{\mathbf{a}^{LM}}$ and RLNM $\mathrm{\mathbf{a}^{RL}}$ to stabilize final mixture action $\mathrm{\mathbf{a}^{R}}$, in which the GoT construction of LFM is designed to evaluate and score the above two executions with more generated evidences or intermediate steps chains from different perspectives.}
\vspace{-10pt}
\label{fig:F4}
\end{figure*}

\subsection{Large Language Feedback Model} 
SALM employs the ability of contextual understanding and inference in LNM to enhance the adaptability of RLNM, aligning and tailoring the pre-trianed DRL policy with different personal preferences and task objectives. Therefore, the integration of DRL-based planner and LM-based planner is facilitated by the LFM, which evaluates both actions and estimates their relative weights as follows:
\begin{equation}
\begin{aligned}
&s_1,s_2 = LFM(\mathbf{a}^{\rm RL},\mathbf{a}^{\rm LM}, \mathbf{a}^{\rm M} ,\mathrm{T}_{\rm HRI} ~ || ~ \mathcal{G}_{GOT})\\
\end{aligned}
\end{equation}
where $\mathbf{a}^{\rm M}$ is a set of executed actions from the memory buffer, and $\mathcal{G}_{GOT}$ is the graph of thought prompting of LFM.

The critical part of LFM is the GoT prompting technique \cite{GoT}. GoT generates and illustrates intermediate reasoning steps as vertices to significantly improve the comprehensibility and inference performance of LFM. Despite requiring additional information and resources, the final inference can be supported more thoroughly with diversified evidence and threads. Moreover, the graph construction also addresses the stochasticity of the inference process through interpretations from multiple reasoning paths. 
%The vertecies of GoT are linked via following three edge transformations (aggregation, refinement, generation, and score), so that the interpretable and editable construction of GoT that can be modified and fine-tuned for different situations.

As shown in Fig.~\ref{fig:F4}, a directed graph framework $\mathcal{G}_{\rm GoT}=\{\mathcal{V},\mathcal{E}\}$ is designed in LFM, in which vertices present solutions in different aspects. Thoughts' transformations or correlations are captured by edges where generation, aggregation, refining, and scoring are typically involved. The aggregation operation is defined as $[ \mathcal{V}^{+}=\{{V}^{+}\}; \mathcal{E}^{+}=\{(V_1,{V}^{+}),\cdots,(V_k,{V}^{+})\} ]$, generation operation is $[ \mathcal{V}^{+}=\{{V}_{1}^{+},\cdots,{V}_{k'}^{+}\}; \mathcal{E}^{+}=\{(V,{V}_{1}^{+}),\cdots,(V,{V}_{k'}^{+})\} ]$, and the refining operation can be presented as $[ \mathcal{V}^{+}=\phi; \mathcal{E}^{+}=\{(V,{V})\} ]$. Additionally, the scoring thought is calculated as $\mathcal{E}(V,\mathcal{G}_{sub},f_{LM})$, where $\mathcal{G}_{sub}$ is a subgraph of $\mathcal{G}_{HRI}$ or the whole graph, and $f_{LM}$ is a pre-trained large model.

The GoT inference procedure is mainly composed by three distinct interactive inference chains, where the additional inference in the intermediate thoughts are constructed for effective comparison and evaluation. Firstly, the GoT process of LFM verify and evaluate the $\rm a^{RL}$, $\rm a^{LM}$, and the pair of ($\rm a^{RL}$, $\rm a^{LM}$), respectively. The prompt engineering of LNM is fed into LFM as input. Meanwhile, the generation operation of GoT of $1^{st}$-layer is introduced to generate more inference or calculation steps, with a checklist text that describe the verification steps (e.g. check current distance between robot and its destination, or calculate next timestep relative distances among robot and humans). Subsequently, $a^{RL}$ and $a^{LM}$ are further evaluated via the first-layer three thoughts with their reasoning evidences to score individual actions as $(s^i_1,s^i_2)$, comparing the effectiveness of different actions. After $2^{nd}$-layer evaluation, the $3^{rd}$-layer vertex is aggregated by two $2^{nd}$-layer thoughts and the $1^{st}$-layer $a^{RL}$ and $a^{LM}$ thoughts to incorporate individual action scores. Finally, the $3^{rd}$-layer vertex generates the relative score thought, which is refined in the $4^{th}$-layer vertex to calculate the combinational scores $(s_1, s_2)$.

\subsection{Reinforcement Learning Navigation Model}

Although LNM adapts the remarkable ability of HRI understanding into navigation decision-making, independent LNM still struggles with uncertainty and the infeasibility of decision-making in complex dynamic environments with continuous space. In contrast, DRL-based RLNM leverages a convergent and efficient policy to address these issues as observed in many works \cite{liu2023intention,wang2023multi, wang2023navistar}. Therefore, inspired by \cite{wang2023navistar}, the DRL-based action execution is employed in SALM with an ST-graph HRI encoder and RLNM block. 

RLNM implicitly models the surrounding long-term environmental dynamics to demonstrate socially acceptable navigation behaviors in human-filled environments, based on a hybrid spatial-temporal transformer $\mathcal{F}_{Trans}$ from current SOTA social navigation benchmark NaviSTAR \cite{wang2023navistar}. Firstly, the underlying human intents and spatial-temporal dependencies are captured by a spatial-temporal transformer framework. Subsequently, the heterogeneous features mentioned above are fused through a multimodal transformer fusion network. Hence, the environmental dynamics of the social navigation scenario, denoted as $ \mathbf X_{\mathrm{E}}$, are constructed as an ST-graph (spatio-temporal graph) $\mathcal{G}_{ST}$ from robot local observations as follows:
\begin{equation}
    \mathbf X_{\mathrm{E}} = \mathcal F_{NaviSTAR} (\mathbf s_1,\cdots,\mathbf s_{\rm{t}} ~ || ~ \mathcal G_{ST}, LLM(Requests))
\end{equation} 

The hybrid spatial-temporal transformer neural network from $ \mathcal F_{NaviSTAR}$ is employed according to \cite{wang2023navistar} as follows:
\begin{equation}
\begin{aligned}
&\mathbf{{X}}_{{\rm S}}^{}, \mathbf{{X}}_{{\rm T}}^{} =  \operatorname{{Trans}_{ST}} 
(\operatorname{Multi}(\{ \mathbf s_1,\cdots,\mathbf s_{\rm{t}} \}))\\
&\mathbf{{X}}_{{\rm E}}^{} =  \operatorname{{Trans}_{F}} 
(\operatorname{CMAtten}(\{\mathbf{{X}}_{{S}}, \mathbf{{X}}_{{T}} \}))
\end{aligned}
\end{equation}
\noindent where $\rm Multi(\cdot)$ denotes the multi-head attention mechanism within the spatial-temporal transformer network $\rm Trans_{ST}(\cdot)$, which maps the sequential observation data into spatial HRI features $\rm \mathbf{X}_S$ and temporal HRI features $ \mathbf{X}_T$. $\rm CMAtten(\cdot)$ represents the cross-attention block in $\rm Trans_{F}$. The cross-attention block not only fuses above spatial and temporal features but also captures their underlying dependencies, enabling an integrated and context representation as follows:
\begin{equation}
\begin{aligned}
        &\operatorname{Atten}\left( \mathbf{{Q}}, \mathbf{{K}}, \mathbf{{V}}\right)={\mathrm{softmax} (\frac{ \mathbf{{Q}} \mathbf ({{\mathbf{K}} })^{\top}}{\sqrt{ {d}_{{h}}}})  \mathbf{{V}} }  
        \\
        &\operatorname{Multi}\left(\mathbf{{Q}} ,\mathbf{{K}} ,\mathbf{{V}} \right)=f_{{fc}}( {Atten}_1,\cdots, {Atten}_{{h}})\\
        &\operatorname{CMAtten}\left(\mathbf{{Q}} ,\mathbf{{K}} ,\mathbf{{V}} \right)= \operatorname{Multi}(\mathbf{{Q}}^{{}}_{{U}}, \mathbf{{K}}^{{}}_{{\bar U}}, \mathbf{{V}}^{{}}_{{\bar U}})
        % &\{\mathbf{{X}}_{{S}}; \mathbf{{X}}_{{T}}\}=  \operatorname{Trans}_{{}}( \operatorname{Multi_S}(\cdot);  \operatorname{Multi_T}(\cdot) | \{\mathbf{O}_{R}, \mathbf{O}_{T}\})\\
\end{aligned}
\end{equation}
\noindent where $d_h$ is the dimension of attention. $\mathbf{{Q}}^{{}}_{{U}}$ and $\mathbf{{K}}^{{}}_{{\bar U}}$, $\mathbf{{V}}^{{}}_{{\bar U}}$ present the different modality query or key-value matrix, where $U \in \{\rm \mathbf{X}_S,~ \rm \mathbf{X}_T\}$ and $\Bar{U}$ is the complemental set of $U$.

%to encode human preference distribution and social norms,

We modify the NaviSTAR \cite{wang2023navistar} planner as our RLNM to address interactive social navigation tasks within the Dec-POSMDP paradigm, in which it generates macro-action $\mathbf{u}^{\rm RL}$ and local-action $\mathbf{a}^{\rm RL}$ based on HRI latent embedding $\mathbf X_{\mathrm{E}}$ as follows:
\begin{equation}
    \mathbf{u}^{\rm RL} , \mathbf{a}^{\rm RL} \sim \mathcal RLNM (\mathbf{X}_{\rm E})
\end{equation} 

% To understand human intents and preferences directly from language, SALM designs an adaptive reward function $r_{LM}$ to enhance the LLM-RLHF training procedure based on \cite{kwon2022reward, ma2023eureka}, which is incorporated with the RLHF reward neural network $r_{\varphi}$ to improve the robustness of the DRL-based action executor. As shown in Fig.~\ref{fig:F5}, the environmental programming specification and personal preference are provided as context prompt engineering to generate the LM reward function corresponding to existing programming formulation and supervisor preference.

% \begin{figure}[!t]
% \centering
% \includegraphics[width=1\linewidth]{F5.pdf}
% \vspace{-20pt}
% \caption{LLM-RLHF block: The LLM is introduced in our RLHF training procedure to effectively interpret human supervision from language feedback, wherein the reward functions are generated directly.}
% \vspace{-10pt}
% \label{fig:F5}
% \end{figure}

%To directly generate LM reward function corresponding to existing programming formulation, the environmental specification is provided as context to LLM. 

%Meanwhile, the context leads to that LLM leverages semantic information of environment variables to construct reward function which is also aligned by requirements and modified by feedbacks. 

Although many DRL-based social navigation approaches \cite{wang2023navistar,wang2023multi,chen2019crowd,liu2023intention} demonstrate excellent performance benchmarks for SAN tasks in latent HRI inference and social compliance collision avoidance, the learning algorithms are limited and constrained by training conditions and biased policy patterns. This limitation makes it difficult to maintain sufficient performance in general application scenarios. Hence, an execution decoder is employed to incorporate action-pair $(\mathbf{a}^{\rm RL}, \mathbf{a}^{\rm LM})$ from both RLNM and LLM, using the relative weights $(\rm s_1, s_2)$ from LFM as follows:
\begin{equation}
\mathbf{a}^{\rm R}_{\rm t}={Decoder}_{}(\rm s_1 \cdot \mathbf{a}^{\rm RL}_{\rm t} + s_2 \cdot \mathbf{a}^{\rm LM}_{\rm t})
\end{equation}
% \noindent where the $Decoder(\cdot)$ is a multilayer perceptron layer that is fine-tuned within the pre-training stage of RLNM.

%Finally, the robot action $\mathbf{a}^{\rm R}_{\rm t}$ is collected into memory block with other state information together.

%Finally, As shown in the DRL-based action execution section of Fig~\ref{fig:F2}, spatio-temporal graph HRI encoder translates human states from robot observations as heterogeneous spatial temporal features to a high-level embeddings as follows:

\section{Experiments And Results}
\subsection{Simulation Experiment}
\subsubsection{Simulation Setup}

\begin{figure}[!t]
\centering
\includegraphics[width=0.55\linewidth]{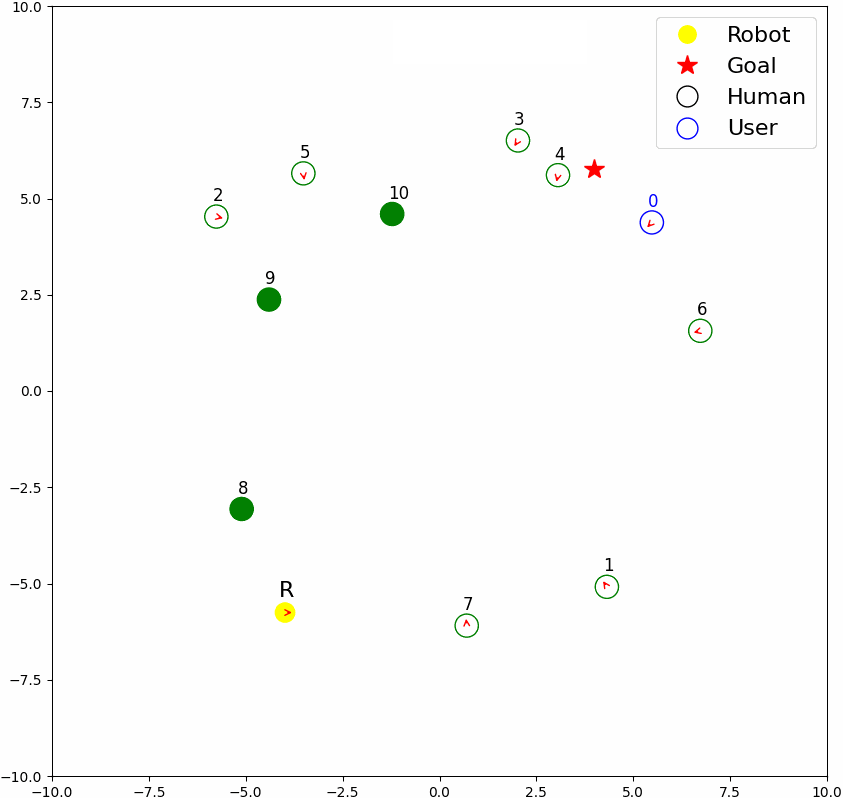}
\vspace{-3pt}
\caption{The illustration of human-in-loop interactive social navigation: The social robot is navigating toward the red star destination with a blue circle user across ten green circle humans. User's feedback is randomly generated with $50\%$ probability to robot(e.g. change the goal position).}
\vspace{-10pt}
\label{fig:env}
\end{figure}

As shown in Fig.~\ref{fig:env}, we conducted simulation experiments to evaluate the performance of our approach and other ablation models. We developed a human-in-loop interactive social robot navigation environment based on a gym social navigation simulator \cite{wang2023navistar, navidiff}. In comparison with the original simulator version, user real-time language commands and feedback can be directly implemented to adjust robot behaviors during execution, where both large model blocks (LLM, LNM, and LFM) are configured by GPT-4o \cite{openai2023gpt4}. However, the kinematics, dynamics, and other environmental constraints remain the same as in previous works \cite{wang2023navistar}.

The default experimental scenario involves a social robot that navigates toward a target in an human-filled open space, while assisting a user to response possible real-time requirements. These human pedestrians each simulated using the ORCA policy \cite{ORCA}. The simulation encompasses two primary interactive task types: human-following (HF) and point-to-point (P2P) navigation tasks. In the HF task, the robot first navigates to meet the user. Once it reaches the user, it continuously updates its destination to remain positioned behind the user, maintaining an appropriate distance until the user arrives at the final destination. For the P2P navigation task, the robot navigates from a starting point to a specified target. Notably, the target may change mid-task based on real-time commands from the user, requiring the robot to adapt its path accordingly.

% All DRL-based models were trained with $1\times10^4$ episodes and tested with 50 random cases for each task, conducted on a desktop with an Intel i9-13900k CPU and an NVIDIA 4090 GPU. In each training or testing epoch, a human language requests generator was designed to publish task objectives with personal preferences. Particularly, real-time feedback is stochastically established from the generator with a $50\%$ probability halfway through the current task to update the robot's goal, user preferences, or other attributes.

\subsubsection{Baselines and Ablation Models}
As shown in Table~\ref{table:result}, we have set up a comparison of our algorithm with five other baselines or ablation models as follows: (1) A traditional navigation strategy ORCA \cite{ORCA} is utilized as the basic planner, wherein only the global LLM block is maintained to understand high-level human commands for task target establishments; (2) We implement CADRL \cite{CADRL} as the baseline for learning-based approaches, in which the LLM block is also employed for interactive navigation; (3) For the second ablation model, LNM and LFM are detached inside to test the performance of RLNM as SA-RLNM; (4) RLNM and LFM are removed to be viewed as the first ablation model, where the robot is driven only from LNM output as SA-LNM; and (5) For the final one, LFM is replaced by a fixed relative parameter pair $(0.5 ~ \& ~ 0.5)$ as SA-LFM.

\subsubsection{Evaluation Metrics}
As shown in Table~\ref{table:result}, all methods have been evaluated using $50$ random test cases within 10 humans and 1 user for each task individually, with two evaluation metrics: successful rate (SR) and social score (SS) \cite{wang2023navistar}. The SS metric considers various social navigation performance factors such as travel time, collision rate, success rate, discomfort level, and etc.

\begin{table}[h]
\vspace{-5pt}
\caption{: Simulation experiment results\label{tab:table1}}
\vspace{-5pt}
\resizebox{\linewidth}{!}{
\centering
\begin{scriptsize}
\begin{tabular}{c|ccc|ccc}
\hline & \multicolumn{3}{c}{ Success Rate } & \multicolumn{3}{c}{Social Score} \\
\cline { 2 - 4 } \cline { 5-7 } Methods & \multicolumn{3}{c}{ Task Type } &  \multicolumn{3}{c}{ Task Type } \\
& P2P & HF & \textcolor{blue}{\textbf{AVG}} & P2P & HF & \textcolor{cyan}{\textbf{AVG}} \\
\hline 
ORCA \cite{ORCA}& $42$ & $20$ & 31 & $35$ & $24$ & 30 \\
CADRL \cite{CADRL}& $64$ & $52$ & 58 & $58$ & $55$  & 57 \\
SRNN \cite{chen2019crowd}& $82$  & $68$ & 75 & $55$  & $53$  & 54 \\
NaviSTAR \cite{wang2023navistar}& $84$ & $64$ & 74 & $70$  & $65$  & 68 \\
SA-RLNM   & $84$ & $66$ & 75 & $83$  & $65$  & 74 \\
SA-LNM & $54$  & $70$ & 62 & $48$  & $58$  & 53 \\
SA-LFM & $64$ & $62$ & 63 & $60$  & $57$  & 59 \\
SALM   & $\mathbf{90}$  & $\mathbf{82}$ & $\mathbf{86}$  & $\mathbf{85}$ & $\mathbf{78}$   & $\mathbf{82}$  \\
\hline
\end{tabular}
\end{scriptsize}
\vspace{-7pt}
\label{table:result}
}
\end{table} 

\subsubsection{Qualitative Analysis}

Table~\ref{table:result} presents a comparative study of our proposed approach alongside five baseline and ablation models, evaluated on two tasks P2P and HF with SR and SS metrics. As a traditional navigation strategy, ORCA utilizes only a global LLM block for high-level human command interpretation. Its relatively low average SR (31) and SS (30) highlight the limitations of classical approaches in complex, socially dynamic environments. For CADRL \cite{CADRL}, this learning-based method, is also struggled on performance with an average SR of 58 and SS of 57.

For the ablation model, SA-RLNM, by detaching the LLM-based LNM and the LFM, this variant relies solely on the DRL-based RLNM. It achieves a higher average SR of 75, yet its social score suggests that while the navigation is effective, the overall social compliance might be compromised. For SR-LNM , both the RLNM and LFM are removed, driving the robot solely from the LNM output. The reduced performance (average SR of 62, SS of 53) underscores the necessity of the DRL-based component for robust navigation. The LFM of SR-LFM  is replaced with fixed parameters, resulting in an average SR of 63 and SS of 59. This suggests that a dynamic, adaptive evaluation mechanism is crucial for optimal performance compared to a static parameter setting. Finally, SALM is integrating the strengths of both LLM-based and DRL-based approaches with an adaptive evaluation model, SALM achieves the highest average SR (86) and SS (82). In summary, the experimental results indicate that our SALM model significantly outperforms both traditional and learning-based baselines. The ablation studies further validate that each component LLM-based navigation, DRL-based planning, and dynamic language feedback—is vital for achieving high performance in socially-aware navigation tasks.

\subsubsection{Effectiveness of LNM}
From Table~\ref{table:result}, the ablation model SA-RLNM achieves a $75\%$ SR and $74$ SS, which is slightly lower than the performance of SALM overall. We can observe that SA-RLNM can provide a benchmark performance derived from  the RLNM capability. However, as shown in the common trajectory maps in the experiment, SA-RLNM demonstrates a similar path quality to SALM in the early timesteps, but SA-RLNM still adheres to previous strategic preferences even after receiving renovation feedback from the user updating the task objective and personal preference. Hence, the introduction of LNM in SALM stabilizes the robustness of navigation performance, especially for the stage after user feedback, because the pre-trained DRL-based policy cannot adjust itself to highly adaptive behavioral representations in the execution stage.

\subsubsection{Effectiveness of RLNM}
Despite the amazing inference ability exhibited by large language models in many applications, LLM-based developments of sequential control systems have struggled with highly dynamic environments and sequential data dimensions. The environments of social navigation tasks require robots to understand potential pedestrian intents and engage in HRI cooperation to adapt to environmental dynamics. Such challenges, coupled with the lack of RLNM compared to SALM, result in struggling and precarious effects in dynamic scenarios. From both the average results and common trajectory maps in the experiment, we observe that SR-LNM exhibits many instances of reciprocal dance phenomena, where the robot swings from side to side. Thus, to maintain benchmarking performance for LLM-based social robots in dynamic spaces, we recommend continued use of DRL-based robot executors.

\subsubsection{Effectiveness of LFM}
As observed in Table~\ref{table:result} and common trajectory maps in the experiment, we find that SR-LFM exhibits limited planner capability compared to SALM. The blunt fusion with fixed relative weights presents lower flexibility than LFM, because the LLM-driven LFM can infer the situation for a better fusion strategy from more evidence chains with the developments of GoT. The blending mechanism of LFM is significant for adjusting and fusing two robot actions, leveraging the inference ability from LLM, which can effectively improve overall performance. More experiment videos can be found at website: \url{https://sites.google.com/view/navi-salm}.

\vspace{-1pt}

\section{Conclusion}
In this work, we developed an interactive social robot large model. SALM leverages the inference ability of LLM to interpret user language commands and  enhance the adaptability of DRL-based navigation policy. Additionally, the GoT is developed by LFM to evaluate the relative action score of the executed actions from LLM-based and DRL-based planners, incorporating LNM and RLNM blocks. Finally, SALM demonstrates outstanding efficiency compared to baselines and ablation models in the experiments. 
%Future work includes exploring its use in real-world robot applications.

%\vspace{-10pt}
\typeout{}
\bibliography{main}
\bibliographystyle{IEEEtran}
\end{document}